\documentclass{article}

\usepackage{arxiv}
\usepackage{authblk}
\usepackage[utf8]{inputenc} 
\usepackage[T1]{fontenc}    
\usepackage{hyperref}       
\usepackage{url}            
\usepackage{booktabs}       
\usepackage{amsfonts}       
\usepackage{nicefrac}       
\usepackage{microtype}      
\usepackage{lipsum}
\usepackage{graphicx}
\usepackage{amsmath}
\usepackage{amssymb}
\usepackage[table,xcdraw]{xcolor}
\usepackage{multirow}
\usepackage{framed,multirow}

\usepackage{amssymb}
\usepackage{latexsym}
\usepackage{amsmath}
\usepackage{url}
\usepackage{xcolor}
\usepackage{float}
\usepackage{color}
\usepackage[normalem]{ulem}

\usepackage{hyperref}

\title{POPCORN: Progressive Pseudo-labeling with Consistency Regularization and Neighboring}

\author[1]{Reda Abdellah Kamraoui}
\author[1]{Vinh-Thong Ta}
\author[2]{Nicolas Papadakis}
\author[1,2]{\\Fanny Compaire}
\author[3]{José V Manjon}
\author[1]{Pierrick Coupé}

\affil[1]{Univ. Bordeaux, Bordeaux INP, CNRS, LaBRI,\\ UMR5800, PICTURA, F-33400 Talence, France}
\affil[2]{Univ. Bordeaux, Bordeaux INP, CNRS, IMB,\\ UMR5251, F-33400 Talence, France}
\affil[3]{ITACA, Universitat Politècnica de València,\\
 46022 Valencia, Spain\\}

\begin{document}
\maketitle
\begin{abstract}
Semi-supervised learning (SSL) uses unlabeled data to compensate for the scarcity of annotated images and the lack of method generalization to unseen domains, two usual problems in medical segmentation tasks. 
In this work, we propose POPCORN, a novel method combining consistency regularization and pseudo-labeling designed for image segmentation. 
The proposed framework uses high-level regularization to constrain our segmentation model to use similar latent features for images with similar segmentations. POPCORN estimates a proximity graph to select data from easiest ones to more difficult ones, in order to ensure accurate pseudo-labeling and to limit confirmation bias.
Applied to multiple sclerosis lesion segmentation, our method demonstrates competitive results compared to other state-of-the-art SSL strategies.
\\ \\ \\

\end{abstract}

\\ \\ \\ \\
\section{Introduction}
Semi-Supervised Learning (SSL) is a promising field which aims to exploit
unlabeled data in order to enhance the performance achieved using only labeled data. SSL is explored to mitigate both problems of the limited availability of labeled data and the lack of model generalization to unseen domains.
Among SSL works proposed for medical image segmentation tasks, we can distinguish three main categories:

\textbf{Consistency Regularization} (CR) constrains the model to give consistent predictions for the same unlabeled input under different perturbations. Bortsova \textit{et al.} \cite{bortsova2019semi} constrained the model to produce similar segmentation when applying different elastic transformations to the same unlabeled images. Similarly, Perone \textit{et al.} \cite{perone2018deep} used a mean teacher strategy where the consistency loss constrained teacher and student predictions to be consistent. Orbes \textit{et al.} \cite{orbes2019multi} designed an adversarial loss to minimize the amount of information for a specific domain and to maximize segmentation consistency. CR offers interesting consistency properties on the learned features but it is usually trained under unrealistic scenarios (e.g., using the same input data under different perturbations). Such oversimplification does not guarantee a good generalization of the learned features.
Besides, some works showed that consistency regularization using perturbation on input data is not adapted for segmentation \cite{frenchsemi,ouali2020semi}.

\textbf{Pseudo-Labeling} (PL) strategies automatically assign labels to unlabeled data in order to use them during training in combination with labeled data. Pseudo-labels are generally assigned by a model trained on labeled data. Uncertainty can be used to measure the confidence of the predictions. For example, Sedai \textit{et al.} \cite{sedai2019uncertainty} employed prediction uncertainty for estimating segmentation confidence on soft labels. Cao \textit{et al.} \cite{cao2020uncertainty} considered an uncertainty aware temporal ensembling strategy. Xia \textit{et al.} \cite{xia20203d} used uncertainty-weighted mechanism for the pseudo-label fusion of multiple networks predictions. PL is a simple way to use unlabeled data. PL is nevertheless prone to confirmation bias (\textit{i.e.}, error propagation) \cite{arazo2020pseudo}. So far, this is the main limitation of PL.

\textbf{Auxiliary Tasks} (AT) are secondary objectives combined with the main segmentation task which do not require ground truth annotations. Using unlabeled data, in such a way, implicitly extracts relevant features for the primary segmentation task. Li \textit{et al.} \cite{li2020shape} proposed the prediction of surface distance maps to capture more effectively shape-aware features. Kervadec \textit{et al.} \cite{kervadec2019curriculum} predicted the size of the target segmentation as an intermediate task. Alternatively, Chen \textit{et al.} \cite{chen2019multi} combined supervised segmentation and unsupervised input reconstruction. Finally, Luo \textit{et al.} \cite{luo2020semi} proposed to predict geometry-aware level set representation of the transformed ground truth annotations. AT demonstrated good performance, but the choice of the AT highly depends on the addressed problem which limits the method generalization for other segmentation tasks.
\\

In this work, our main contribution is threefold:
\begin{itemize}
\item We propose a novel framework that combines consistency regularization and pseudo-labeling for segmentation.
\item We propose a consistency regularization strategy that ensures proximity in latent space of images with similar segmentations. This allows us to produce meaningful feature representation and accurate predictions. 
\item We propose a new pseudo-labeling strategy which selects progressively unlabeled samples according to their similarity with training data, in order to limit confirmation bias.

\end{itemize}

\section{Method}
\subsection{Method overview}
\begin{figure*}[t]
    \centering
    \newcommand{\sz}{1\textwidth}
    \includegraphics[width=\sz]{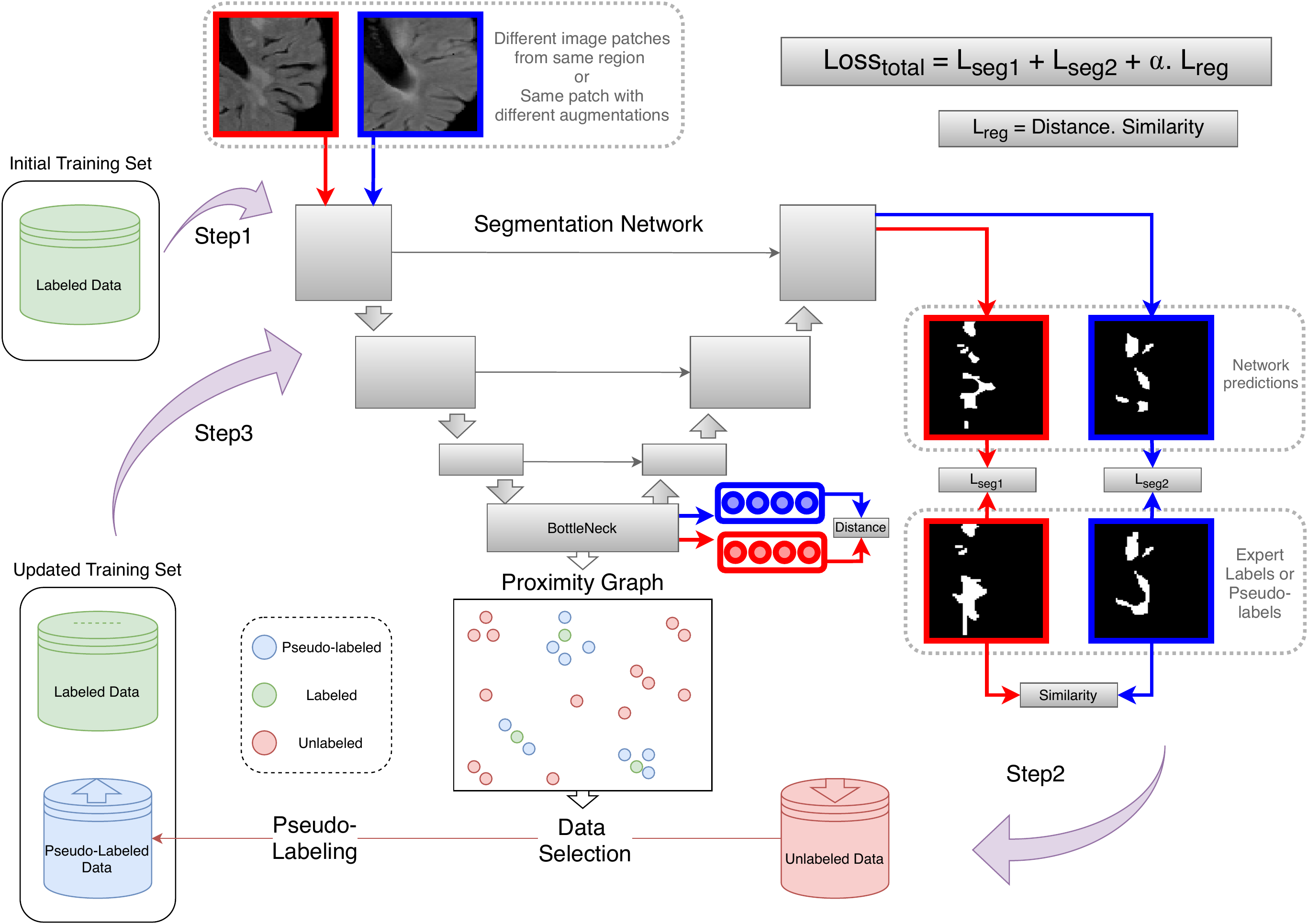}
      \caption{The training process of POPCORN}
    \label{fig:pipline}
\end{figure*}

The proposed strategy is a PrOgressive Pseudo-labeling with COnsistency Regularization and Neighboring (POPCORN) for semi-supervised learning in segmentation (see Fig.\ref{fig:pipline}).
First, the training is performed with a new CR ensuring that: $i)$ augmented versions of the same image have identical feature maps, and $ii)$ images with similar segmentations have similar feature maps.
Second, PL of the unlabeled data is performed gradually. At each selection step, the proximity graph is used to select new unlabeled samples. The pseudo-labels of the chosen data-points are estimated with the current segmentation model and incorporated in the training set. 

The main intuition is that our segmentation model is able to produce more accurate pseudo-labels for images similar to our training set. Since our CR ensures close features for similar data, features extracted from the model are used to select new samples. 

\subsection{Bottleneck consistency regularization} \label{reg}
In POPCORN, the model architecture is based on 3D U-Net composed of an encoder and a decoder, linked by a bottleneck and skip connections at different scales (see Fig.\ref{fig:pipline}).
For an image $X$,  $F(X)$ represents the prediction of the segmentation network, and $h(X)$ represents the latent features of $X$ extracted at the bottleneck level.
Our method is based on a dual/hybrid loss ensuring segmentation quality and consistency relevance.
\subsubsection{Segmentation loss:}
As traditionally done in supervised learning, we use the Dice similarity loss as the first element of our global loss. This loss ensures the similarity of the produced output with the expected one.

\begin{equation}
L_{seg}( F(X), y) = 1 - Dice( F(X), y), 
\end{equation}
where $y$ is either the expert segmentation of $X$ when available, or the pseudo-label otherwise.

\subsubsection{Consistency regularization loss:}
Alongside $L_{seg}$, a regularization loss on the bottleneck is used:
\begin{equation}
L_{reg} [\, (X_i,y_i), (X_j,y_j) \, ] = Distance( \,h( X_i ),\, h( X_j )\, )\times Similarity( y_i , y_j ), 
\end{equation}
where $X_i$, $X_j$ are image patches randomly selected as either different augmented versions of the same patch, or patches from different images extracted from the same region and with the same orientation.
Moreover, let us define:
\begin{equation}
Similarity( y_i , y_j ) = e^ { - mse(y_{i},y_{j})}, 
\end{equation}
\begin{equation}
Distance( h_i, h_j ) = \frac{2\, ||h_i-h_j||^2}{||h_i||^2+||h_j||^2}, 
\end{equation}
where $mse$ is the mean squared error, and $||\,.\,||$ is the euclidean distance.

The total loss is a combination of $(1)$ and $(2)$ with a weighting coefficient $\alpha$:
\begin{equation}\label{loss_total}
L_{total}  [ (X_i,y_i), (X_j,y_j) ] = L_{seg}( F(X_i), y_i) +  L_{seg}( F(X_j), y_j)\\ + \alpha. L_{reg} [ (X_i,y_i), (X_j,y_j) ] \,.
\end{equation}

\subsection{Pseudo-labeling data selection} \label{pseudo}
Curriculum learning \cite{bengio2009curriculum} showed that presenting data with an increasing difficulty can lead to a better learning process. We consider unlabeled data close to the training data as easy examples to be incorporated first in the training process, whereas distant samples are considered more challenging.
Indeed, the latent distance between unlabeled and training data can be viewed as a measure of similarity. Thus, pseudo-labeling using the trained model is more accurate for unlabeled samples which are similar to training data.

Our data selection is performed in three steps (see Fig.\ref{fig:pipline}).
Step1: the training set is limited to labeled data. Once the segmentation model is trained until convergence, it is used to extract latent space representation for each unlabeled datapoint.
Step2: the proximity graph is used to select $K$ unlabeled data that guarantee a smooth learning (as described in \ref{graph}). For each selected unlabeled datapoint, a pseudo-label (segmentation) is assigned by the trained model.
Step3: the model is trained for $N$ epochs with the new training set containing both labeled and pseudo-labeled data.
Step two and three are repeated every $N$ epochs by picking each time $K$ new data points, their respective pseudo-labels are being computed with the newly updated segmentation model. The process is maintained until all unlabeled data are integrated into the training set.

\subsection{Proximity graph}\label{graph}

The proximity graph represents the euclidean distance between the training and the unlabeled samples latent representations:
\begin{equation}
P_{i\in U,j\in T}= ||h(X_i)-h(X_j)||^2
\end{equation}
where $T$ and $U$ represent respectively the training set (labeled and pseudo-labeled), and unlabeled data.
For data selection, we propose the following criteria to select $K$ elements of $U$ close to $T$.
For each datapoint of $U$, the proximity with $T$ is defined by the sum of the $p$ closest elements of $T$ to the datapoint:
\begin{equation}
[i^{\star}_1,i^{\star}_2…,i^{\star}_K] = Argmin^K( \sum_{j\in Argmin^p(P_i)}   P_{i,j} ), 
\end{equation}
where $[i^{\star}_1,i^{\star}_2…,i^{\star}_K]$ represent the $K$ indices of the data selected. $Argmin^k(V)$ returns the indices of the $k$ smallest values of the vector $V$.

\section{Experiments}

\subsection{Dataset}

\, \, \, \textbf{Labeled Data:} 
For labeled data, the ISBI training dataset \cite{carass2017longitudinal} is used. It consists of 21 longitudinal multimodal images (including FLAIR modality) from only five different subjects with Multiple Sclerosis (MS). The images have been acquired on the same MRI scanner. MS lesions were delineated by two expert raters. 
This dataset has limited image quality diversity (all the images were acquired with the same protocol on a single site) and inter-subject variability (only 5 subjects).

\textbf{Unlabeled Data:}
The unlabeled dataset consists of 2901 FLAIR MRI (large inter-subject variability) with white matter hyperintensities. It does not only contain MS, which increases pathology diversity. MRI have been collected across multiple acquisition sites based on different manufacturers, 1.5T and 3T scanners, 2D and 3D sequences. 
This dataset covers a large diversity in terms of image quality, pathology and inter-subject variability. 

\textbf{Testing Data:}
For assessing our results, the dataset described in \cite{coupe2018lesionbrain} is used. It contains 3D multimodal MRI from 43 subjects diagnosed with MS. The images have been acquired with three different scanners and different acquisition protocols. Consequently this dataset proposes a larger diversity than the labeled dataset. Lesion masks have been obtained by expert manual delineation.

All images have been pre-processed using the same pipeline \cite{coupe2018lesionbrain}.

\subsection{Reference Methods}\label{reference}
POPCORN is compared to state-of-the-art strategies \cite{chen2019multi}, \cite{sedai2019uncertainty} and \cite{bortsova2019semi}. The following strategies have been implemented based on their published works and adapted to MS lesion segmentation. 
First, the multi-task attention-based SSL \cite{chen2019multi} is an AT strategy. It combines supervised segmentation and unsupervised reconstruction objectives. The reconstruction task uses an attention mechanism to predict input image regions of different classes.
Second, the uncertainty guided pseudo-labeling \cite{sedai2019uncertainty} is a PL strategy. The teacher model, trained only on labeled data, generates soft segmentation (pseudo-labels) and uncertainty maps for all the unlabeled data at once. The uncertainty is used for estimating segmentation confidence of the generated segmentation when training the student model.
Finally, the semi-supervised transformation consistency \cite{bortsova2019semi} is based on CR. In addition to the primary loss, a consistency loss ensures that the prediction of the same images under transformations are consistent.

\subsection{Implementation details}
The method hyperparameters were chosen empirically according to the size of labeled and unlabeled datasets.
First, 200 from the $M=2901$ unlabeled images were chosen after each training cycle that ran for 2 epochs ($K=200,\, N=2$) to limit computational burden.
Second, the number of neighbors $p=5$ was selected considering the initial training data of 21 labeled images. We suggest that this value is a good compromise in order to consider relevant near neighbors while avoiding far neighbors which mislead data selection.

In addition, we used the architecture proposed by \cite{kamraoui2020towards} with a patch size of $[64\times64\times64]$ and a threshold of 0.5 to obtain the binary segmentation.
Moreover, image quality data augmentation was used to introduce realistic perturbations, where blur, edge enhancement, and other augmentations simulated image quality heterogeneity \cite{kamraoui2020towards}. 
Furthermore, the coefficients for the regularization part of the loss have been set to 0.2 ($ \alpha = 0.2$).
Finally, the experiments have been performed with Keras 2.2.4 \cite{chollet2015keras} and Tensorflow 1.12.0 \cite{abadi2016tensorflow} on Python 3.6. The model was optimized with Adam \cite{kingma2014adam} using a learning rate of 0.0001 and a momentum of 0.9.

\subsection{Statistical Analysis}
To assert the advantage of a technique obtaining the highest average score, we conducted a Wilcoxon test over the lists of Dice scores measured at image level. The significance of the test is established for a p-value below 0.05.

\section{Results} 
\subsection{Ablation study} \label{ablation}
To evaluate our contributions, we compare POPCORN with other versions of our strategy when isolating key elements.
As shown in Table \ref{tab:ablation}, our full method achieves the highest Dice and the second best result in terms of precision.
First, when comparing POPCORN without consistency regularization (corresponds to $\alpha=0$ in \eqref{loss_total}) and our full method, we notice a decrease in both precision and sensitivity. This suggests that without CR, the latent space is less meaningful for our selection process of unlabeled data.
Second, to underline the impact of the proximity graph, we consider another progressive PL strategy where pseudo labels are randomly selected. Although the strategy without proximity graph is slightly more sensitive, we observe an important drop in both Dice and precision compared to our full method. This demonstrates that the proposed progressive selection based on image proximity in latent space is more robust to confirmation bias than random selection.
Next, when running only half the selection steps $(M=1400)$, our method obtained the second best Dice score. This shows that POPCORN with nearly half unlabeled data can achieve better performance than the other variations and methods with full dataset (see also \ref{results}). 
Finally, when combining the proposed CR (on labeled data only) with the baseline (supervised learning), the precision is considerably improved. This shows the importance of our CR on segmentation accuracy, beyond data selection. Overall, the statistical analysis shows that our full method has a significantly higher Dice than the baseline, the version without CR, baseline with CR, and Ours without proximity graph.

\begin{table}[t]
\centering
\caption{
The table details the impact of each contribution: the consistency regularization (CR), the proximity graph, and using labeled/pseudo-labeled (lab/pseudo) data. Best result is displayed in bold, and the second best result is underlined.}

\resizebox{\textwidth}{!}{%
\begin{tabular}{c|c|c|c|c|c}
Strategy & Trained on & CR on  & Dice    & Precision       & Sensitivity     \\ \hline
\rowcolor{gray!20} Our method    & \begin{tabular}[c]{@{}c@{}}Lab + Pseudo\end{tabular} & \begin{tabular}[c]{@{}c@{}}Lab + Pseudo\end{tabular} & \textbf{73,09\%} & \underline{73,33\%} & 74,29\% \\

 Ours with half selection steps ($M=1400$)  & \begin{tabular}[c]{@{}c@{}}Lab + Pseudo\end{tabular} & \begin{tabular}[c]{@{}c@{}}Lab + Pseudo\end{tabular} & \underline{70,59\%} & 68,26\% & \textbf{75,91\%} \\

\rowcolor{gray!20} \begin{tabular}[c]{@{}c@{}}Ours without CR\end{tabular}   & \begin{tabular}[c]{@{}c@{}}Lab + Pseudo\end{tabular} & None & 69,13\% & 70,49\% & 70,58\% \\

Ours without proximity graph  & \begin{tabular}[c]{@{}c@{}}Lab + Pseudo\end{tabular} &  \begin{tabular}[c]{@{}c@{}}Lab + Pseudo\end{tabular} & 68,06\% & 65,14\%          & \underline{74,40\%} \\

\rowcolor{gray!20} \begin{tabular}[c]{@{}c@{}}Baseline with CR\end{tabular}     & Lab                                                            & Lab                                                             & 68,08\%          & \textbf{77,77\%}          & 61,94\%          \\

Baseline                                                                    & Lab                                                             & None                                                                & 64,41\%          & 61,80\%          & 69,70\%         
\end{tabular}%
}

\label{tab:ablation}
\end{table}

\subsection{Comparison with state-of-the-art approaches} \label{results}

Table \ref{tab:results_strategies} shows the results of POPCORN compared to the reference methods presented in section \ref{reference}.
First, all the SSL strategies obtain a significantly better Dice scores compared to the baseline.
Second, POPCORN obtains the highest Dice followed by Uncertainty guided Pseudo-labeling \cite{sedai2019uncertainty}. 
Next, the multi-task attention-based SSL \cite{chen2019multi} and the semi-supervised transformation consistency \cite{bortsova2019semi} respectively obtain the highest precision and sensitivity rates.
Finally, POPCORN obtains the best balance between precision and sensitivity, as opposed to the other strategies which are more prone to FP \cite{bortsova2019semi,sedai2019uncertainty} and FN \cite{chen2019multi}.
Overall, POPCORN has a significantly higher Dice compared to the other methods according to our Wilcoxon test.  

\begin{table}[htbp]
\centering
\caption{The table represents results of POPCORN (our method) compared to other state-of-the-art strategies on the testing dataset (see \ref{results} for complementary details).}
\resizebox{400pt}{!}{%
\begin{tabular}{c|c|c|c}
Strategy                                      & Dice            & Precision             & Sensitivity             \\ \hline
\rowcolor{gray!20}POPCORN                                       & \textbf{73,09\%} & \underline{73,33\%}          & \underline{74,29\%} \\
 Multi-task Attention-based SSL \cite{chen2019multi}    & 67,23\%          & \textbf{75,72\%} & 61,99\%          \\
\rowcolor{gray!20}Uncertainty guided Pseudo-labeling \cite{sedai2019uncertainty}          & \underline{68,31\%}          & 67,93\% & 71,95\%          \\
 Semi-supervised transformation consistency \cite{bortsova2019semi} & 66,75\%          & 61,52\%          & \textbf{78,79\%} \\
\rowcolor{gray!20}Baseline (labeled data only)                                    & 64,41\%          & 61,80\%          & 69,70\%          
\end{tabular}%
}

\label{tab:results_strategies}
\end{table}

\begin{figure*}[t]
    \centering
    \newcommand{\sz}{1\textwidth}
    \includegraphics[width=\sz]{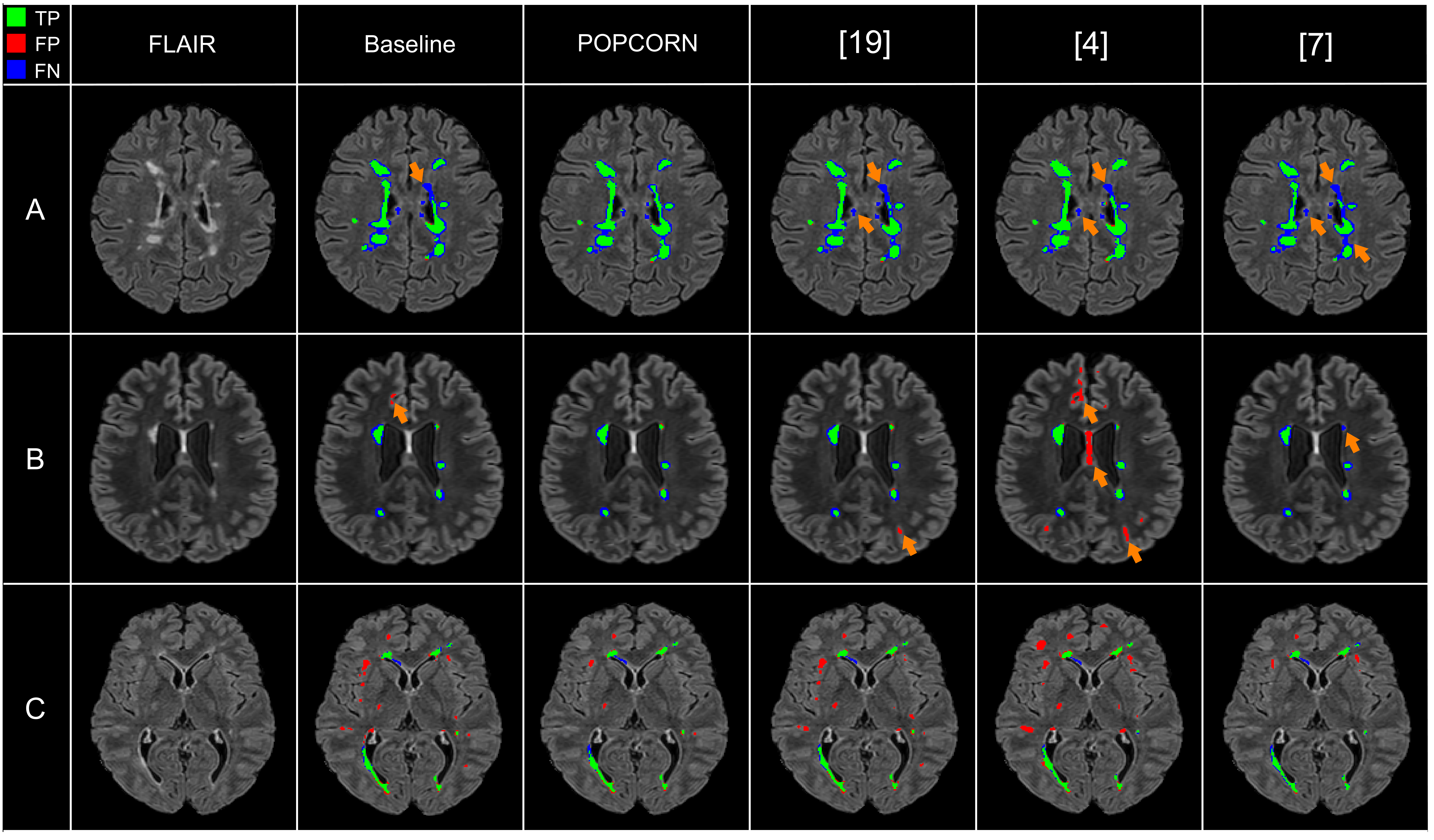}
      \caption{Comparison of POPCORN, Uncertainty guided Pseudo-labeling \cite{sedai2019uncertainty}, multi-task attention-based SSL \cite{chen2019multi}, and the semi-supervised transformation consistency \cite{bortsova2019semi} lesion segmentations. Orange arrows indicate key segmentation differences. }
    \label{fig:sample}
\end{figure*}

Fig.\ref{fig:sample} shows image segmentations produced by POPCORN and the compared strategies. A, B, and C are images from the testing dataset, specifically chosen to showcase acquisition and lesion diversity. 
For A, we observe that POPCORN segmentation is the most accurate. On the contrary, \cite{chen2019multi,sedai2019uncertainty} are the least sensitive with high volumes of false negative. Similarly, the segmentations obtained with the baseline and \cite{bortsova2019semi} do not cover all lesions.
On image B, the segmentation provided by \cite{bortsova2019semi} contains several false positive lesions, compared to the other strategies. Both the baseline and \cite{sedai2019uncertainty} only include one or two false detections. POPCORN proposes an accurate segmentation. Last, the method \cite{chen2019multi} misses a small lesion. 
For C, we notice that \cite{bortsova2019semi,sedai2019uncertainty} and the baseline detect many false positive lesions. POPCORN and \cite{chen2019multi} produce fewer false detection on this challenging sample.
To conclude, our strategy segments accurately most lesions while minimizing false detection. Compared to the other strategies, POPCORN maintains the best balance between the sensitivity and the precision of lesion segmentation.

\section{Conclusion}
We propose a novel strategy for SSL segmentation. Our method combines consistency regularization and pseudo-labeling.  POPCORN progressively selects unlabeled samples with an increasing difficulty using a proximity graph. Overall, we have shown the improvement of using POPCORN compared to other state-of-the-art strategies, as well as the impact of each of our contributions.

\section{Acknowledgements}
This work benefited from the support of the project DeepvolBrain of the French National Research Agency (ANR-18-CE45-0013). This study was achieved within the context of the Laboratory of Excellence TRAIL ANR-10-LABX-57 for the BigDataBrain project. Moreover, we thank the Investments for the future Program IdEx Bordeaux (ANR-10-IDEX-03-02, HL-MRI Project), Cluster of excellence CPU and the CNRS/INSERM for the DeepMultiBrain project. This study has been also supported by the DPI2017-87743-R grant from the Spanish Ministerio de Economia, Industria Competitividad. The authors gratefully acknowledge the support of NVIDIA Corporation with their donation of the TITAN Xp GPU used in this research.

\bibliographystyle{abbrv}
\bibliography{paper.bbl}  


\end{document}